\documentclass[journal]{IEEEtran}
\ifCLASSINFOpdf
\else
\fi
\usepackage{amsmath}
\usepackage{algorithmic}
\usepackage{array}

\usepackage{amsmath,amsthm,amssymb,epsf}
\usepackage{bm} %% Bold Math
\usepackage{graphicx,psfrag}
\usepackage{epstopdf}
\usepackage{verbatim}
\usepackage{multirow}
\usepackage{color}
\usepackage{setspace}
\usepackage{enumitem}
\usepackage{subfigure}
\usepackage{amssymb}
\usepackage{algorithm}
\usepackage{algorithmic}
\usepackage{tabularx}
\usepackage{booktabs}
\usepackage{csquotes}
\newcommand*{\Scale}[2][4]{\scalebox{#1}{$#2$}}%
\begin{document}
\title{Assured Learning-enabled Autonomy: A Metacognitive Reinforcement Learning Framework}

\author{Aquib~Mustafa,  Majid~Mazouchi, Subramanya Nageshrao% <-this % stops a space
       , Hamidreza~Modares,~\IEEEmembership{Senior Member,~IEEE,}
\thanks{This work was supported by Ford Motor Company-Michigan State University Alliance.}
\thanks{A. Mustafa, M. Mazouchi, and H. Modares are with the Department of Mechanical Engineering, Michigan State University, East Lansing, MI 48824 USA (e-mail: mustaf15@msu.edu; mazouchi@msu.edu; modaresh@msu.edu).}% <-this % stops a space
\thanks{S. Nageshrao is with Ford Research and Innovation Center, Ford Motor Company, Palo Alto, CA 94304, USA (snageshr@ford.com).}}


\markboth{ }%
{Mustafa \MakeLowercase{\textit{et al.}}: Assured Learning-enabled Autonomy: A Metacognitive Reinforcement Learning Framework}

\maketitle

\begin{abstract}
Reinforcement learning (RL) agents with pre-specified reward functions cannot provide guaranteed safety across variety of circumstances that an uncertain system might encounter. To guarantee performance while assuring satisfaction of safety constraints across variety of circumstances, an assured autonomous control framework is presented in this paper by empowering RL algorithms with metacognitive learning capabilities. More specifically, adapting the reward function parameters of the RL agent is performed in a metacognitive decision-making layer to assure the feasibility of RL agent. That is, to assure that the learned policy by the RL agent satisfies safety constraints specified by signal temporal logic while achieving as much performance as possible. The metacognitive layer monitors any possible future safety violation under the actions of the RL agent and employs a higher-layer Bayesian RL algorithm to proactively adapt the reward function for the lower-layer RL agent. To minimize the higher-layer Bayesian RL intervention, a fitness function is leveraged by the metacognitive layer as a metric to evaluate success of the lower-layer RL agent in satisfaction of safety and liveness specifications, and the higher-layer Bayesian RL intervenes only if there is a risk of lower-layer RL failure. Finally, a simulation example is provided to validate the effectiveness of the proposed approach.
\end{abstract}


\IEEEpeerreviewmaketitle

\section{Introduction}

\IEEEPARstart{R}{einforcement} learning (RL) is a goal-oriented learning approach, inspired by biological systems, and is concerned with designing agents that can take actions in an environment so as to maximize some notion of cumulative reward \cite{Sutton:1}-\cite{Rizvi:010}.  Despite tremendous success of RL in a variety of applications, including robotics \cite{Kober:5}, control \cite{Sutton:6}, and human-computer interaction \cite{Chavarriaga:7}, existing results are categorized as weak artificial intelligence (AI) \cite{Lu:8}. That is, current RL practice has mainly been used to achieve pre-specified goals in structured environments by handcrafting a cost or reward function for which its minimization guarantees reaching the goal. Strong AI, on the other hand, holds the promise of designing agents that can learn to achieve goals across multiple circumstances by generalizing to unforeseen and novel situations. As the designer cannot foresee all the circumstances that the agent might encounter, pre-specifying and handcrafting the reward function cannot guarantee reaching goals in an uncertain and non-stationary environment. Reward shaping \cite{Ng:9}-\cite{Konidaris:10} has been presented in the literature with the primary goal of speeding up the learning without changing the outcome of the solution. Intrinsic motivated RL \cite{Chentanez:11} has also been presented so that agents learn to shape their reward function to better trade-off between exploration and exploitation or learn faster for the applications that the external environmental reward is sparse. 

In control community, several RL-based feedback controllers have been presented for control of uncertain dynamical systems \cite{Kiumarsi:12}-\cite{Cheng:16}. In these traditional RL-based controllers, the reinforcement signal feedback is derived through a fixed quadratic objective function \cite{Lewis:17}-\cite{Si:18}. A fixed reward or objective function, however, cannot guarantee achieving desired specifications across all circumstances. To express rich specifications rather than quadratic objectives, temporal logic, as an expressive language close to human language, has been widely used. It is natural to leverage temporal logic for specifying goals and incorporating domain knowledge for the RL problem \cite{Liu:19}-\cite{Saha:22}. RL with temporal logic specifications has also been used recently \cite{Li:23}-\cite{Sun:027}. However, defining the rewards solely based on temporal logic specifications and ignoring numerical rewards can result in sparse feedbacks in control systems and cannot include other performance objectives such as energy and time minimization. Moreover, the system pursues several objectives and as the circumstance changes, the system's needs and priorities also change, requiring adapting the reward signal to encode these needs and priorities to the context. It is therefore desired to design a controller that provides a good enough performance across variety of circumstances while assuring that its safety-related temporal logic specifications are satisfied. 

To this end, this paper takes a step towards strong AI for feedback control design by presenting a notion of adaptive reward function by introducing a metacognitive layer that decides on what reward function to optimize depending on the circumstance. More specifically, a metacognitive assured RL framework is presented to learn control solutions with good performances while satisfying desired specifications and properties expressed in terms of Signal temporal logic (STL). We first discuss that RL agents with pre-specified reward functions cannot guarantee satisfaction of the desired specified constraints and performances across all circumstances that an uncertain system might encounter. That is, the system either violates safety specifications or achieves no optimality and liveness specifications. To overcome this issue, a metacognitive decision-making layer is augmented with the RL agent to learn what reward functions to choose to satisfy desired specifications and to achieve a good enough performance across variety of circumstances. More specifically, a fitness function is defined in the metacognitive layer that indicates how safe the system would react in the future for a given reward function and in case of a drop in the fitness function, a Bayesian RL algorithm will proactively adapt the reward function parameters to maximize system's assuredness (i.e., satisfaction of the desired STL safety and liveness specifications) and guarantee performance. Off-policy RL algorithms are proposed to find optimal policies corresponding to each hyperparameter by reusing the data collected from the system. The proposed approach separates learning the reward function that satisfies specifications from learning the control policy that maximizes the reward and thus allows us to evaluate as many hyperparameters as required using reused data collected from the system dynamics.

{\bf Outline}: The paper is organized as follows: Section II provides the used notations and preliminaries in the paper. Section III outlines the motivation and formulates the problem statement. Section IV presents the metacognitive monitoring and control framework. Section V presents a low-level RL-based control architecture. The simulation results are stated in Section VI. Finally, concluding remarks are provided in Section VII.

\section{NOTATIONS AND PRELIMINARIES}

\subsection{Notations}
Throughout the paper, $\mathbb{R}$ and  $\mathbb{N}$ represent the sets of real numbers and natural numbers, respectively. $\mathbb{R}^n$ denotes $n$-dimensional Euclidean space. The superscript ${(.)^T}$ denotes transposition. $I$ denotes the identity matrix of proper dimension. ${[K]_j}$ denotes the $j$-th element of the vector  $K$. ${[K]_{ij}}$ denotes the $\left[ {i,j} \right]$-th entry of the matrix $K$. $diag(A)$ denotes a diagonal matrix in which all off-diagonal entries are zero, i.e., ${[A]_{ij}} = 0,\,\,\forall i \ne j$. ${\rm{Tr}}(A)$ stands for trace of the matrix $A$. When a random variable ${\varepsilon _i}$ is distributed normally with mean $m$ and variance ${w^2}$, we use the notation ${\varepsilon _i} \sim  {\cal N}(m, {w^2})$.   $\otimes$ denotes Kronecker product and $vec\left( A \right)$ denotes the $mn$-vector constructed by stacking the columns of matrix $A \in \mathbb{R}{^{n \times m}}$ on top of one another.

\smallskip

\noindent
{\bf Definition 1.} Let $X$ and $Z$ be two sequences with probability distributions $P_X$ and $P_Z$, respectively.
The weighted Kullback–Leibler (KL) divergence between $X$ and $Z$ is defined as \cite{McIntire:27}
\begin{align}
D_{KL}^h(X||Z) = \int {{P_X}(\theta )\log {{\left( {\frac{{{P_X}(\theta )}}{{{P_Z}(\theta )}}} \right)}^{h(x)}}d\theta }  \label{eq:1}
\end{align}
where $h(x)$  is a non-negative real-valued weighting function. 

Note that the weighting function is defined to weigh more heavily promising regions of the state space. Note also that ${D_{KL}}({P_X}||{P_Z}) \ge 0$  and  ${D_{KL}}({P_X}||{P_Z}) = 0$ if and only if ${P_X} = {P_Z}$. 

\subsection{Signal Temporal Logic}

Temporal logic can be used to express rich time-dependent specifications or constraints for a broad range of control system applications. Signal temporal logic (STL) is a category of temporal logic, which enables to specify temporal properties for real-valued and continuous-time signals \cite{Donze:28}-\cite{Lindemann:31}. Let $x(t)$ be a continuous-time signal. The STL predicates $\sigma $  are evaluated as True $( \top )$  or False $( \bot )$  according to a corresponding predicate function ${z^\sigma }(x):\mathbb{R}^n \to \mathbb{R}$ as
\begin{align}
\sigma  = \left\{ \begin{array}{l}
 \top \,\,\,\,{z^\sigma }(x) > 0\\
 \bot \,\,\,\,{z^\sigma }(x) \le 0
\end{array} \right. \label{eq:2}
\end{align}

The predicate function ${z^\sigma }(x)$ is a linear or nonlinear combination of the elements of the signal $x$  and the predicate $\sigma $ belongs to a set of predicates ${P_\sigma } = \{{\sigma _1},{\sigma _2}, \ldots ,{\sigma _N}\}$ with $N \in \mathbb{N}$ denoting the number of predicates. One can recursively combine predicates using Boolean logic \textit{negation} ($\neg$), \textit{disjunction} ($\vee$) and \textit{conjunction} ($\wedge$) as well as temporal operators \textit{eventually} ($\diamondsuit$), globally or \textit{always} ($\square$) and \textit{until} ($U$) to form complex formulas $\varphi$, known as task specifications, as
\begin{align}
\varphi : =  \top |\sigma |\neg \sigma |\,\,{\varphi _1} \wedge {\varphi _2}|{\varphi _1} \vee {\varphi _2}|{\diamondsuit _{[a,b]}}\varphi |\,{\square _{[a,b]}}\varphi |\,\,{\varphi_1}\,{U_{[a,b]}}{\varphi_2}. \nonumber
\end{align}

%The time bounds of the \textit{until} operator $\varphi \,{U_{[a,b]}}\mu$ are given as $a,b \in [0,\infty )$ with $a< b$. The commonly used temporal operators \textit{eventually} and \textit{always} follow from ${\diamondsuit _{[a,b]}}\varphi  =  \top {U_{[a,b]}}\varphi $, respectively. For example, the temporal formula ${\diamondsuit _{[3,6]}}\varphi$ is satisfied when the STL formula $\varphi$ becomes True within the time interval of $3$ to $6$ seconds.

For each predicate ${\sigma _i},\,\,i = 1,...,N$, a predicate function ${z^\sigma }(x(t))$ is defined as (2). The following qualitative semantics are used to express that a signal $x(t)$ satisfies an STL expression at the time $t$ \cite{McIntire:27}-\cite{Lindemann:30}.
\begin{align}
\begin{array}{l}
(x,t) \models \sigma \,\,\,\,\,\,\,\,\,\,\,\,\,\,\,\,\,\,\,\,\,\, \Leftrightarrow \,\,\,{z^\sigma }(x(t)) > 0\\
(x,t) \models \neg \sigma \,\,\,\,\,\,\,\,\,\,\,\,\,\,\,\,\,\, \Leftrightarrow \,\,\,\neg ((x,t) \models \sigma )\\
(x,t) \models \varphi  \wedge \mu \,\,\,\,\,\,\,\,\,\,\,\, \Leftrightarrow \,\,\,(x,t) \models \varphi  \wedge (x,t) \models \mu \\
(x,t) \models \varphi  \vee \mu \,\,\,\,\,\,\,\,\,\,\,\, \Leftrightarrow \,\,\,(x,t) \models \varphi  \vee (x,t) \models \mu \\
(x,t) \models \varphi {U_{[a,b]}}\mu \,\,\,\,\,\, \Leftrightarrow \,\,\,\exists {t_1} \in [t + a,\,\,t + b]\,\,s.t.\,\,(x,{t_1}) \models \mu \\
\qquad\qquad\qquad\quad\quad \qquad  \wedge \,\,\, \forall {t_2} \in [t,\,\,t_1]\,\,s.t.\,\,(x,{t_2}) \models \varphi\\
(x,t) \models {\diamondsuit _{[a,b]}}\varphi \,\,\,\,\,\,\,\,\, \Leftrightarrow \,\,\,\exists {t_1} \in [t + a,\,\,t + b]\,\,s.t.\,\,(x,{t_1}) \models \varphi \\
(x,t) \models {\square_{[a,b]}}\varphi \,\,\,\,\,\,\,\,\, \Leftrightarrow \,\,\,\forall {t_1} \in [t + a,\,\,t + b]\,\,s.t.\,\,(x,{t_1}) \models \varphi. 
\end{array} \label{eq:3}
\end{align}

The symbol  $\models$ denotes satisfaction of an STL formula. The time interval $[a,b]$ differentiates STL from general temporal logic and defines the quantitative timing to achieve the continuing temporal formula. 

Apart from syntaxes and qualitative semantics, STL provides also various robustness measures to quantify the extent to which a temporal constraint is satisfied. Given STL formulas $\varphi$ and $\mu $, the spatial robustness is defined as [30]
\begin{align}
\begin{array}{l}
{\rho ^\sigma }(x,t) = {z^\sigma }(x(t))\\
{\rho ^{\neg \sigma }}(x,t) =  - {\rho ^\sigma }(x,t)\\
{\rho ^{\varphi  \wedge \mu }}(x,t) = \min ({\rho ^\varphi }(x,t),{\rho ^\mu }(x,t))\\
{\rho ^{\varphi  \vee \mu }}(x,t) = \max ({\rho ^\varphi }(x,t),{\rho ^\mu }(x,t))\\
{\rho ^{\varphi {U_{[a,b]}}\mu }}(x,t) = \mathop {\max }\limits_{{t_1} \in [t + a,\,t + b]} (\min ({\rho ^\mu }(x,{t_1}),\mathop {\min }\limits_{{t_2} \in [t,\,{t_1}]} ({\rho ^\varphi }(x,{t_2}))\\
{\rho ^{{\diamondsuit _{[a,b]}}\varphi }}(x,t) = \mathop {\max }\limits_{{t_1} \in [t + a,\,t + b]} {\rho ^\varphi }(x,{t_1})\\
{\rho ^{{\square_{[a,b]}}\varphi }}(x,t) = \mathop {\min }\limits_{{t_1} \in [t + a,\,t + b]} {\rho ^\varphi }(x,{t_1}).
\end{array} \label{eq:4}
\end{align}

This robustness measure determines how well a given signal $x(t)$ satisfies a specification. The space robustness defines such a real-valued function ${\rho ^\sigma }(x,t)$ which is positive if and only if $(x,t) \models \sigma$. That is, ${\rho ^\sigma }(x,t) > 0\,\, \Leftrightarrow \,\,(x,t) \models \sigma$. Let a trajectory $\tau [0,T]$ be defined by the signals $x(t)$ throughout its evolution from time $0$ to $T$. A trajectory then satisfies the specification if and only if ${\rho ^\sigma }(x,t) > 0\,\,\forall \,t \in [0,{t_f}]$, where ${t_f}$  is the end time of the STL horizon. 

\subsection{Gaussian Process}

A Gaussian process (GP) can be viewed as a distribution over functions, in the sense that a draw from a GP is a function. GP has been widely used as a nonparametric regression approach for approximating a nonlinear map $f:X \to \mathbb{R}$ from a state $x$ to the function value $f(x)$. The function values $f(x)$ are random variables and any finite number of them has a joint Gaussian distribution. When a process $f$ follows a Gaussian process model, then
\begin{align}
f(.) \sim  GP({m_0}(.), {k_0}(.\,,.)) \label{eq:5}
\end{align}
where ${m_0}(.)$ is the mean function and ${k_0}(.,.)$ is the real-valued positive definite covariance kernel function \cite{Rasmussen:32}. In GP inference, the posterior mean and covariance of a function value $f(x)$ at an arbitrary state $x$ is obtained by conditioning the GP distribution of $f$ on past measurements. Let ${X_n} = [{x_1},...,{x_n}]$ be a set of discrete state measurements, providing the set of inducing inputs. For each measurement ${x_i}$, there is an observed output ${y_i} = f({x_i}) = m({x_i}) + {\varepsilon _i}$ where ${\varepsilon _i} \sim  {\cal N}(0, {w^2})$. The stack outputs give $y = [{y_1},...,{y_n}]$.  The posterior distribution at a query point $x$ is also a Gaussian distribution and is given by \cite{Rasmussen:32}
\begin{align}
{m_n}(x) &= {m_0}(x) + K{(x,{X_n})^T}{({K_n} + I\,{w^2})^{ - 1}} \nonumber \\
{k_n}(x,x') &= {k_0}(x,x') - K{(x,{X_n})^T}{({K_n} + I\,{w^2})^{ - 1}}K(x,{X_n})
 \label{eq:6}
\end{align}
where the vector $K(x,{X_n}) = [{k_0}(x,{x_1}),...,{k_0}(x,{x_n})]$ contains the covariance between the new data, $x$, and the states in ${X_n}$, and ${[{K_n}]_{ij}} = {k_0}({x_i},{x_j}),\,\,\forall i,j\, \in \{ 1,...,n\}$.

\section{PROBLEM STATEMENT AND MOTIVATION}

In this section, the problem of optimal control of systems subject to the desired specifications is formulated. We then discuss that optimizing a single reward or performance function cannot work for all circumstances and it is essential to adapt the reward function to the context to provide a good enough performance and assure safety and liveness of the system. 

Consider the non-linear continuous-time system given by
\begin{align}
\dot x(t) = f(x(t)) + g(x(t))u(t) \label{eq:7}
\end{align}
where $x \in X$ and $u \in U$ denote the admissible set of states and inputs, respectively. We assume that $f(0) = 0$, $f(x(t))$ and $g(x(t))$ are locally Lipschitz functions on a set $\Omega  \subseteq {\mathbb{R}^n}$ that contains the origin, and that the system is stabilizable on $\Omega $. 

The control objective is to design the control signal $u$ for the system (7) to 1) make the system achieve desired behaviors (e.g., track the desired trajectory ${x_d}(t)$  with good transient response) while  2) guaranteeing safety specifications specified by STL, i.e., guaranteeing $(x,t) \models \sigma$ where $\sigma $ belongs to a set of predicates ${P_\sigma } = [{\sigma _1},{\sigma _2}, \ldots ,{\sigma _N}]$  with $N$  as the number of the constraints. To achieve these goals, one can use an objective function for which its minimization subject to $(x,t) \models \sigma$ provides an optimal and safe control solution aligned with the intention of the designer. That is, the following optimal control formulation can be used to achieve an optimal performance (encoded in the reward function $r(.)$) while guaranteeing STL specifications. 

\smallskip

\noindent
{\bf Problem 1 (Safety-Certified Optimal Control).} Given the system (7), find a control policy $u$ that solves the following safe optimal control problem. 
\begin{align}
\begin{array}{l}
\min J(x(t),u(t),{x_d}(t)) = \int_t^\infty  {{e^{ - \gamma (\tau - t )}}r(x(\tau ),u(\tau ),{x_d}(\tau ))} \,d\tau \\
{\rm{s}}{\rm{.t}}{\rm{.}}\,\,(x,\tau ) \models \sigma ,\,\,\,\forall \tau  \ge t\,\,
\end{array} \label{eq:8}
\end{align}
where $\gamma $ is a positive discount factor, 
\begin{align}
r(x(t),u(t),{x_d}(t)) = \sum\limits_j {{q_j}} \,{r_j}(x(t),u(t),{x_d}(t)) \label{eq:9}
\end{align}
is the overall reward function with ${r_j}(x(t),u(t),{x_d}(t))$ as the cost for the $j$-th sub-goal of the system with ${q_j}$ as its weight function, and ${x_d}(t)$ is an external signal (e.g., reference trajectory). 

The optimization framework in Problem 1, if efficiently solved, works well for systems operating in structured environments in which the system is supposed to perform a single task and the priorities across sub-goals (i.e., ${q_j}$ in (9)) do not change over time and reference trajectories need not be adjusted. However, first of all, Problem 1 is hard to solve as it considers both optimality and safety in one framework. Second, even if efficiently solved, for complex systems such as self-driving cars for which the system might encounter numerous circumstances, a fixed reward function cannot capture the complexity of the semantics of complex tasks across all circumstances. As the circumstance changes, the previously rewarding maneuvers might not be achievable safely and thus the feasibility of the solution to Problem 1 might be jeopardized. 

\smallskip

\noindent
{\bf Definition 2. (Feasible Control Policy).} Consider the system (7) with specifications $\sigma $.  A control policy $u(t) = \mu (x)$ is said to be a feasible solution to Problem 1 if 
\begin{enumerate}
\item[{1)}] $\mu (x)$ stabilizes the system (7) on $\Omega$. 
\item[{2)}] There exists a safe set $S \in X$ such that for every ${x_0} \in S$, ${x_t}({x_0},\mu ) \models \sigma, \,\,\forall t$, where ${x_t}({x_0},\mu )$ is the state trajectory $x(t)$ and time $t \ge 0$ generated by (7) with the initial condition ${x_0}$ and the policy $u = \mu (x)$. 
\item[{3)}] $J(x,u,{x_d}) \le \infty $ for all $x \in \Omega$. 
\end{enumerate}

The feasibility of Problem 1 can be jeopardized as the context changes unless the reward weights and/or the reference trajectory are adapted to the context. For example, consider the case where a vehicle is performing some maneuvers with a desired velocity safely under a normal road condition. If, however, the friction of the road changes and the vehicle does not adapt its aspiration towards its desired reference trajectory, when solving Problem 1, it must either violate its safety specifications or the performance function will become unbounded as the system's state cannot follow the desired speed without violating safety. Since the performance will be unbound for any safe policy, the vehicle might only wander around and not reach any goal, providing very poor performance.  This highlights the importance of proposing a metacognitive framework that adapts to the context. 

\smallskip

\noindent
{\bf Remark 1.} One might argue that the original weights or desired reference trajectory in Problem 1 can be appropriately designed in a context-dependent fashion to ensure satisfaction of the desired specifications across variety of circumstances. However, during the design stage, it is generally not possible to foresee the circumstances that will cause violation of the desired specifications and come up with a context-dependent reward function. This is generally due to modeling errors, unknown changes in the environment, and operator intervention. 

Solving Problem 1 for systems with uncertain dynamics is hard. While RL algorithms can solve optimal control problems for systems with uncertain dynamics, they typically do so without taking into account safety constraints. To deal with this challenge, in this paper, we use two layers of control to solve Problem 1 and guarantee its feasibility. In the lower layer, an RL algorithm is used to find an optimal controller that minimizes the performance (8) without considering the safety constraints. The metacognitive layer then monitors safety constraints and their level of satisfaction to proactively make meta-decisions about what reward function to optimize to guarantee feasibility of Problem 1 as the context changes. To guarantee satisfaction of the desired specifications with maximum assuredness, the metacognitive layer must be added on top of the lower-layer optimal control design to decide about priorities over sub-goals as well as the adaptation of the desired reference trajectory. The metacognitive layer monitors the system-level operations and provides corrective action by optimizing a fitness function that guarantees systems' liveness and safety, and thus ensures the maximum assuredness across different circumstances. 

\section{METACOGNITIVE CONTROL ARCHITECTURE}

To find an optimal solution while always guaranteeing satisfaction of the desired specifications with maximum assuredness, as shown in Fig. 1, a metacognitive RL algorithm is presented and it consists of:
\begin{itemize}
  \item A low-level RL-based controller ${\cal K}$ for the system ${\cal S}$ minimizes the performance (8) without considering the safety constraints. 
  \item A high-level metacognitive controller ${\cal C}$ adapts the reward function for the low-level controller ${\cal K}$ to guarantee feasibility of Problem 1 and to maximize assuredness. 
\end{itemize}

The goal is to synthesize the controller ${\cal C}$ for the system (1) to assure that the closed-loop system achieves the desired objective encoded in minimization of a low-level cost function $J(x(t),u(t),{x_d}(t))$ in (8) while guaranteeing systems' liveness and safety, i.e., $(x,t) \models \sigma$ in the metacognitive layer. Separation of RL control design to optimize the performance and metacognitive design to maximize assuredness by optimizing a fitness function significantly simplifies solving Problem 1 and allows to present data-based techniques for solving it.  Let ${\theta _1} \in {\mathbb{R}^{{d_1}}}$, ${\theta _2} \in {\mathbb{R}^{{d_2}}}$, and ${\theta _3} \in {\mathbb{R}^{{d_3}}}$ be vectors of parameters in matrices $Q({\theta _1})$, $R({\theta _2})$, and $x_d^{}({\theta _3})$. Let $\mathchar'26\mkern-10mu\lambda $ be defined as the set of all admissible hyperparameters $\theta $ where $\theta : = {[\theta _1^T,\theta _2^T,\theta _3^T]^T}$. Note that we assume that the set of all admissible parameters $\theta  \in \mathchar'26\mkern-10mu\lambda $ is predefined by the designer based on some prior knowledge. With a slight abuse of notation, we write ${r_\theta }$, ${Q_\theta }$, and ${R_\theta }$ instead of $x_d^{}({\theta _3})$, $Q({\theta _1})$, and $R({\theta _2})$ in what follows. 
 
 \begin{figure}[!t]
 \centering{\includegraphics[width=2.5in] {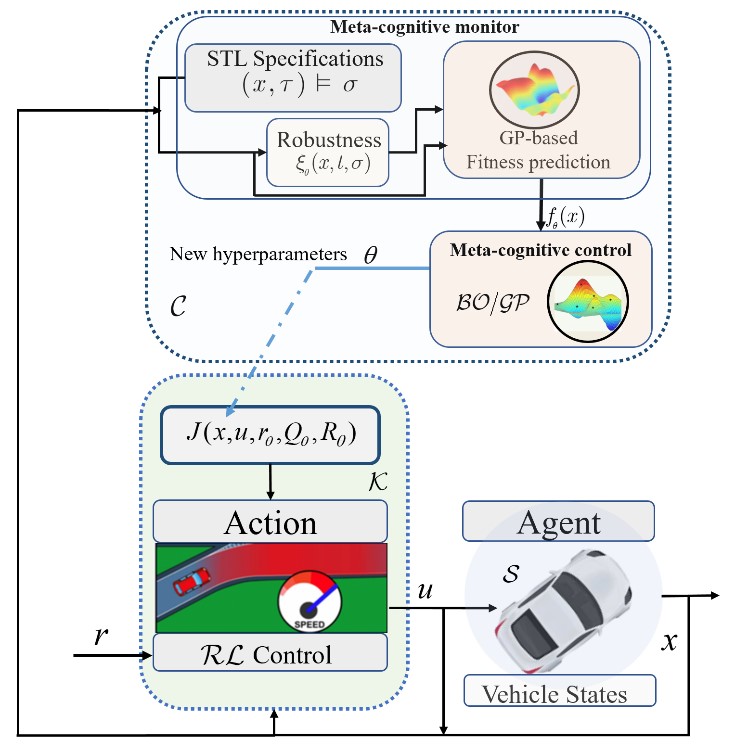}}
 \caption{Proposed metacognitive control scheme. ${\cal S}$: The system to be controlled; ${\cal K}$: low-level RL controller; ${\cal C}$: high-level metacognitive layer scheme.}
 \label{fig1}
 \end{figure}

\subsection{Metacognitive layer Monitoring and Control }

The focus of this subsection is on designing a metacognitive layer that guarantees feasibility of Problem 1, regardless of the type of the RL-based controller (e.g., policy gradient, policy iteration, etc) used in the lower layer. While the proposed approach is not limited to any specific type of performance, in the sequel, we consider an optimal setpoint tracking problem with STL safety specifications. That is, we consider the following parameterized reward function (9) in terms of the hyperparameter vector $\theta $.
\begin{align}
r(x(\tau ),u(\tau ),r(\tau ),\theta ) =& {(x(\tau ) - {r_\theta })^T}{Q_\theta }(x(\tau ) - {r_\theta }) \nonumber \\
 &+ \,{u^T}(\tau ){R_\theta }u(\tau ) \label{eq:10}
\end{align}
where ${Q_\theta }$ and ${R_\theta }$ are the parametrized design weight matrices, which are assumed diagonal, and ${r_\theta }$ is the desired setpoint. The hyperparameter can be then defined as the stack of all parameters of the design weight matrices and the desired setpoint. The performance function in Problem 1 then becomes
\begin{align}
J(x,u,r) =& \int_t^\infty  {{e^{ - \gamma (\tau - t)}}[{{(x(\tau ) - {r_\theta })}^T}{Q_\theta }(x(\tau ) - {r_\theta })\,} \nonumber \\
 &+ \,{u^T}(\tau ){R_\theta }u(\tau ))]d\tau  \label{eq:11}
\end{align}

\smallskip

\noindent
{\bf Assumption 1.} Let $u_\theta ^*(x)$ be the optimal solution to Problem 1 for a set of hyperparameters $\theta $ in the performance function. At each circumstance, there exists a $\theta $ such that its corresponding optimal control policy $u_\theta ^*(x)$ is a feasible solution to Problem 1. 

\smallskip

\noindent
{\bf  Remark 2.} Note that the lower layer receives the hyperparameter vector $\theta $ from the metacognitive layer and any RL algorithm for an unknown system can be used to minimize the accumulated reward function for the specified $\theta $ and determine the optimal control policy in the lower layer. While the presented framework considers a model-free RL-based control approach in the lower layer, it can be applied to any performance-driven-control architecture such as \cite{Lewis:33}-\cite{Jiang:34}. 

The metacognitive layer monitors the functionality of the system-level operations and performs corrective actions for the lower-level controller when required. That is, hyperparameters (i.e., design weights and setpoint) are adjusted in the metacognitive layer to guarantee feasibility of Problem 1 with maximum assuredness as the context changes. To monitor the functionality of the lower-layer, i.e., to check if it performs as intended, accumulated robustness degree of temporal logic specifications is used to define a fitness function as a metric for measuring the system's safety (constraint satisfying) and liveness (goal-reaching and reference tracking). If the fitness function drops, which is an indication that the feasibility of the solution to Problem 1 is about to be violated, the metacognitive layer adapts the hyperparameters of the reward function to guarantee feasibility. Since feasibility of Problem 1 is only guaranteed if the safety STL specifications are satisfied and the performance remains bounded, we use degree of robustness of the safety STL specifications as well as degree of robustness of goal or reference reaching STL specifications (liveness specifications to assure the performance boundness) and define a fitness as a metric to proactively monitor the feasibility of Problem 1 and reacts before STL specifications are violated. 

\subsection{Metacognitive Monitoring}

While $N$ safety specifications are considered as constraints in Problem 1, we define $(N + 1)$-th specification as the liveness (setpoint tracking) of the system as the following STL formula.
\begin{align}
{\sigma _{N + 1}} = {\diamondsuit _{[0,{t_s}]}}\neg (\left\| {x(t) - {r_\theta }} \right\| > \varepsilon ) \label{eq:12}
\end{align}
with $\varepsilon $ as the envelope on the tracking error, and ${t_s}$ as the expected settling time. 

\smallskip

\noindent
{\bf Lemma 1.} Let $u_\theta ^*(x)$ be the optimal control policy found by solving Problem 1 with the performance (11). If $u_\theta ^*(x)$ is feasible, then {${x_t}({x_0},u_\theta ^*) \models {\sigma _{N + 1}}, \forall t$}. 

\smallskip

\noindent
{\bf Proof.} For a given $\theta $, under the assumption that there exists a stabilizing control policy, it is shown in \cite{Lewis:33} that the performance is bounded for the optimal controller $u_\theta ^*(x)$. On the other hand, based on Barbalat lemma \cite{Khalil:35}, a uniformly continuous real function, whose integral up to infinity exists and is bounded, vanishes at infinity. Therefore, the performance is bounded for $u_\theta ^*(x)$ if it makes $\left\| {x(t) - r(t)} \right\|$ become very small after the settling time ${t_s}$. This completes the proof. \hfill $\square$

\smallskip

\noindent
{\bf Remark 3.} Incorporating goal-reaching specifications can help adapt the reward function to avoid performing unintended functionalities in many applications. A classic example for which this could have helped resolve the problem is OpenAI's demo (https://openai.com/blog/faulty-reward-functions/) for which an RL agent in a boat racing game kept going in circles while repeatedly hitting the same reward targets to gain a high score without having to finish the course.

The STL specification $(x,t) \models {\sigma _{N + 1}}$ essentially states that the trajectory tracking error should eventually become less than $\varepsilon$ after the expected settling time. Otherwise, the setpoint $r$ is aimed too high to be achieved safely and must be adjusted. The settling time could be obtained from the knowledge that we have from the control expectation from the lower layer and can be conservative. We now extend the set of predicates from $P = [{\sigma _1},...,{\sigma _N}]$  with $N$ as the number of the constraints to  $P = [{\sigma _1},...,{\sigma _N},{\sigma _{N + 1}}]$ to include the liveness STL predicate. The monitor then predicts if $(x,t) \models \sigma$ will be satisfied all the time to make proactive meta-decisions accordingly. Let the stack of predicate functions for safety and liveness specification STL ${\sigma _i} \in {P_\sigma }$ be 
\begin{align}
{z^\sigma }(x) = \left[ {\begin{array}{*{20}{c}}
{{z^{{\sigma _1}}}(x(t))}\\
{{z^{{\sigma _2}}}(x(t))}\\
 \vdots \\
{{z^{{\sigma _{N + 1}}}}(x(t))}
\end{array}} \right]. \label{eq:13}
\end{align}

Based on (12), the predicate function for liveness becomes
\begin{align}
{z^{{\sigma _{N + 1}}}}(x(t)) = \varepsilon  - \left\| {x(t) - {r_\theta }} \right\|. \label{eq:14}
\end{align}

Using ${z^\sigma }(x)$, a fitness function is now designed to monitor and estimate the accumulated level of satisfaction of the desired STL specifications (i.e., the safety-value function) in the metacognitive layer. If the fitness function drops, which is an indication that either the safety constraints are about to be violated in the future or liveness of the system will not guaranteed, i.e., the feasibility of the solution to Problem 1 is about to be violated, then metacognitive layer proactively adapts the hyperparameters of the reward function to guarantee feasibility.

\smallskip

\noindent
{\bf Definition 3.} Consider a specific hyperparameter vector $\theta $  and let $u = \mu (x)$ be a feasible solution to Problem 1 with the performance (11). The set $S \in X$  is called a viable set (safe and live) of  $\mu (x)$ if for every ${x_0} \in S$, ${x_t}({x_0},\mu ) \models \sigma, \forall t$ and is defined as
\begin{align}
{S_{\mu ,\theta }}(x) = \{ {x_0}:{x_t}({x_0},\mu ) \models \sigma ,\,\,\forall t \ge 0\}  \label{eq:15}
\end{align}
where $\sigma $ belongs to the set of predicates $P = [{\sigma _1},...,{\sigma _N},{\sigma _{N + 1}}]$ with the predicate functions given in (13) .

Note that the dependence of ${S_{\mu ,\theta }}(x)$ on $\theta $ is because the specification (14) depends on $\theta $.

\smallskip

\noindent
{\bf Lemma 2.} Consider a specific hyperparameter vector $\theta $. If the set ${S_{\mu ,\theta }}(x)$ is empty for all control policies $\mu (x) \in U$, then there exists no feasible solution to Problem 1 for the hyperparameter vector $\theta $. 

\smallskip

\noindent
{\bf Proof.} The set of predicates $P$  includes all safety constraints with predicates  ${\sigma _1},...,{\sigma _N}$ as well as the liveness condition with predicate ${\sigma _{N + 1}}$ defined in (12). If the set ${S_{\mu ,\theta }}(x)$  is empty, then, there is no control policy $\mu (x)$ to simultaneously satisfy all safety constraints and make the performance bounded (based on Lemma 1). Therefore, based on Definition 1, there is no feasible solution to Problem 1. This completes the proof. \hfill $\square$

To monitor the feasibility of Problem 1 for the current hyperparameter vector $\theta $, we now define a fitness function based on the quantitative semantics of STL specifications under an optimal control policy found by minimizing (11) for  the given $\theta $. Let $u_\theta ^*(x)$ be the optimal control policy found by minimizing the performance function (11) for a given $\theta $  and is applied to the system. I will be shown in the next section how to use off-policy learning to find optimal solutions for many hyperparameters while only a behavior policy is applied to the system to collect data. Based on (4), once $u_\theta ^*(x)$ is applied to the system, to monitor the degree of robustness of specifications under $u_\theta ^*(x)$, one can write the conjunction over predicate functions $P = [{\sigma _1},...,{\sigma _N},{\sigma _{N + 1}}]$ as
\begin{align}
\begin{array}{l}
{\xi _\theta }(x,t,\sigma ) =  \wedge _{i = 1}^{N + 1}{\rho ^{{\sigma _i}}}(x,t) = \\
\mathop {min}\limits_{i \in [1, \ldots ,N + 1]} \left( {{\rho ^{{\sigma _1}}}(x,t),...,{\rho ^{{\sigma _{N + 1}}}}(x,t))} \right)
\end{array} \label{eq:16}
\end{align}
where ${\rho ^{{\sigma _i}}}(x,t) = {z^{{\sigma _i}}}(x(t))$  and $x(t) = {x_t}({x_0},u_\theta ^*(x))$ is the state trajectory $x(t)$ and time $t \ge 0$  generated by (7) with the initial condition ${x_0}$  and the policy $u_\theta ^*(x)$.

In order to avoid the non-smooth analysis, a smooth under approximation for ${\xi _\theta }(x,t,\sigma )$ is provided in the following lemma.  

\smallskip

\noindent
{\bf Lemma 3.} Consider conjunction of $N + 1$ predicate functions given in (13) and their overall robustness $\xi (x,t,\sigma )$ defined in (16). Then, 
\begin{align}
\xi _\theta ^a(x,t,\sigma ) \buildrel \Delta \over =  - \sum\limits_{i = 1}^{N + 1} {\log \,[{e^{ - {\rho ^{{\sigma _i}}}(x,t)}}} ] \le {\xi _\theta }(x,t,\sigma ) \label{eq:17}
\end{align}

\smallskip

\noindent
{\bf Proof.} See \cite{Lindemann:31}. \hfill $\square$

\smallskip

\noindent
{\bf Lemma 4.} The sign of the function $\xi _\theta ^a(x,t,\sigma )$ is the same as the sign of the function  ${\xi _\theta }(x,t,\sigma )$.

\smallskip

\noindent
{\bf Proof.}   It is immediate from (17) that  ${\xi _\theta }(x,t,\sigma ) < 0$  results in $\xi _\theta ^a(x,t,\sigma ) < 0$.  On the other hand, if ${\xi _\theta }(x,t,\sigma ) > 0$, then based on (16), ${\rho ^{{\sigma _i}}}(x,t) > 0$  for all $i \in [1,...,N + 1]$  and thus $ - \log ({e^{ - {\rho ^{{\sigma _i}}}(x,t)}}) > 0,\,\,\,i = [1,...,N + 1]$, which completes the proof.   \hfill $\square$                                                                                 

Now, a fitness function for a hyperparameter vector $\theta $  as a metric for measuring the system’s safety and liveness in terms of overall robustness is defined as 
\begin{align}
{f_\theta }(x(t)) =& \int_t^\infty  {{e^{ - a(\tau  - t)}}(1 - l)[\log (\xi _\theta ^a\,(x,\tau ,\sigma ))}  + \nonumber \\
&(1 + l)\log (1 + \xi _\theta ^a{\,^{ - 1}}(x,\tau ,\sigma ))]d\tau  \label{eq:18}
\end{align}
where $l = {\mathop{\rm sgn}} (\xi _\theta ^a(x,\tau ,\sigma ))$. The first term is a barrier function to make the fitness infinity if the degree of robustness becomes negative (STL specifications are violated). On the other hand, the lower the fitness, the better the robustness of safety and liveness specifications. This is because the inverse of the degree of robustness is used in the fitness function. 

Note that for the nonempty set ${S_{\mu ,\theta }}(x)$  in (15), the fitness function in (18) becomes
\begin{align}
{f_\theta }(x(t)) = \int_t^\infty  {{e^{ - a(\tau  - t)}}[} 2\log (1 + \xi _\theta ^a{\,^{ - 1}}(x,\tau ,\sigma ))]d\tau  \label{eq:19}
\end{align}

\smallskip

\noindent
{\bf Theorem 1.} There exists a control policy $\mu (x)$ for which ${S_{\mu ,\theta }}(x)$ is nonempty, if and only if the fitness function in (18) is bounded over some set. 

\smallskip

\noindent
{\bf Proof.} If the set ${S_{\mu ,\theta }}(x)$ is empty for all $\mu (x) \in U$, then, for any initial condition ${x_0}$ and any one has ${x_t}({x_0},\mu ) \mathrel{\setbox0=\hbox{$\models$}
 \rlap{\hbox to\wd0{\hss/\hss}}\box0} \sigma $ and consequently  $\xi _\theta ^{}(x,t,\sigma ) < 0  \Rightarrow \xi _\theta ^a(x,t,\sigma ) < 0$  for some time  $t$. This makes the fitness function (18) unbounded because of the first term. On the other hand, if the set ${S_{\mu ,\theta }}(x)$ is nonempty, then for some control policy $\mu (x) \in U$ and any initial condition in  ${x_0} \in {S_{\mu ,\theta }}(x)$, ${x_t}({x_0},\mu ) \models \sigma ,\,\,\forall t \ge 0$. Thus, $\xi _\theta ^{}(x,t,\sigma ) > 0, \,\forall t \ge 0$. Based on Lemma 4, $\xi _\theta ^a(x,t,\sigma ) > 0, \,\forall t \ge 0$.   Let ${\varepsilon _0} = \mathop {\min }\limits_{t \ge 0} \{ \xi _\theta ^a(x,t,\sigma )\}  < \infty $. Then, 
\begin{align}
{f_\theta }({x_0}) &\le 2\log (1 + \frac{1}{{{\varepsilon _0}}})\int_t^\infty  {{e^{ - a(\tau  - t)}}} \,d\tau  
 \nonumber \\
&\le \frac{2}{a}\log (1 + \frac{1}{{{\varepsilon _0}}}) < \infty,  \label{eq:20}
\end{align}
$\forall {x_0} \in {S_{\mu ,\theta }}(x)$. This completes the proof. \hfill $\square$

We now present an online data-based approach to learn the fitness function as a function of the state. Define the meta-reward as
\begin{align}
\begin{array}{l}
{r_m}(x,t) = (1 - l)\log (\xi _\theta ^a\,(x,t,\sigma )) + \\
\,\,\,\,\,\,\,\,\,\,\,\,\,\,\,\,\,\,\,(1 + l)\log (1 + \xi _\theta ^a{\,^{ - 1}}(x,t,\sigma ))
\end{array} \label{eq:21}
\end{align}

The fitness function corresponding to $u_\theta ^*(x)$ at one specific state can now be interpreted as the accumulated meta-rewards the system receives starting from that state when  $u_\theta ^*(x)$ is applied to the system, and thus it can be interpreted as a safety-related value for that state. To calculate the fitness function for all states in a set of interest, one can run the system from all those states to collect a trajectory and then calculate the fitness function. This, however, is not practical and not data efficient. To obviate this issue, the fitness function in  (18) is written as
\begin{align}
{f_\theta }(x(t)) = \int_t^{t + T} {{e^{ - a(\tau  - t)}}} {r_m}(x,\tau )\,\,d\tau  + {f_\theta }(x(t + T)) \label{eq:22}
\end{align}
where ${f_\theta }(x(t + T))$  is the fitness value sampled at the time $t + T$ for a given $\theta$. This equation resembles the Bellman equation in RL and its importance is that it allows us to express the fitness values of states as fitness values of other sampled states. This opens the door for value-function approximation-like approaches with function approximation for calculating the fitness value for each state. That is,  since the fitness of consecutive samples are related using (22), a parametrized form of the fitness function or a nonparametric form of it can be used to  learn the fitness function for all states of interest using only a single trajectory of the system. This will allow fast proactive decision-making in the upper layer and will prevent the system from reaching an irreversible crisis, for which no action can keep the system in its safety envelope in the future. To this end, a data-based approach will assess the fitness online in real-time without using a model of the system.  Once the fitness is learned, a monitor will detect changes in the fitness and consequently the situation.\par
\vspace{-0.07cm}
We first consider learning the fitness function. Since the form of the fitness function is not known, Gaussian processes (GP) is employed to estimate the function ${f_\theta }(x(t))$. In analogy to GP regression for RL \cite{Engel:36}, a GP prior is first imposed over the fitness function, i.e., 
\begin{align}
{f_\theta }(x) \sim  {\cal N}\Big({m_0}(x,\theta), {k_0}\big((x,\theta),(x',\theta')\big)\Big) \nonumber
\end{align}
with a mean ${m_0}(x,\theta)$ and covariance ${k_0}\big((x,\theta),(x',\theta')\big)={k_x(x,x') \otimes k_{\theta}(\theta,\theta')}$. Note that $k_x(x,x')$ captures similarity in the state space and $k_{\theta}(\theta,\theta')$ captures similarity in the space of reward functions. The parameter $\theta$ can also include preview information to also reflect similarity in the context provided by auxiliary information that can be measured (e.g. road curvature, traffic situation, etc) (i.e, $k_{\theta}(\theta,\theta')={k_{\theta_a}(\theta_a,\theta_a') \otimes k_{\theta_b}(\theta_b,\theta_b')}$ where the first term correspond to reward parameters and the second term correspond to the preview information). This will provide transferring across contexts and will significantly reduce the frequency of updating the higher-layer GP model. To employ GP regression, based on (22), the temporal difference (TD) error for fitness function is written as 
\begin{align}
{f_\theta }(x(t)) - {f_\theta }(x(t + T)) = \int_t^{t + T} {{e^{ - a(\tau  - t)}}} {r_m}(x,\tau )d\tau \label{eq:23}
\end{align}

To learn the fitness function using GP, for a given $\theta$, the sequence of samples or a trajectory ${x_1},...,{x_L}$ generated from applying its corresponding control policy is used to present the following generative model.
\begin{align}
R({x_{t + T}}) = {f_\theta }(x(t)) - {f_\theta }(x(t + T)) + \delta (t) \label{eq:24}
\end{align}
where
\vspace{-0.15cm}
\begin{align}
R({x_i}) = \int_{{t_{i - 1}}}^{{t_i}} {{e^{ - a(\tau  - {t_{i - 1}})}}} {r_m}(x,\tau )\,\,d\tau  \nonumber
\end{align}
and $\delta (t) \sim  {\cal N}(0,{w^2})$ denotes zero-mean Gaussian noise indicating uncertainty on the fitness function. Note that (24) can be considered as a latent variable model in which the fitness function plays the role of the latent or hidden variable while the meta-reward plays the role of the observable output variable. As a Bayesian method, GP computes a predictive posterior over the latent values by conditioning on observed meta-rewards. Let $X_L^{^\theta } = [{x_1},...,{x_L}]$  with ${x_t} = {x_t}({x_0},u_\theta ^*(x))$ be the trajectory collected after $u_{{\theta _i}}^*(x)$ (the optimal control input found by the lower-layer RL for the hyperparameter vector $\theta $) applied to the system. An algorithm for the derivation of $u_\theta ^*(t)$, for the new hyperparameter vector  $\theta $ based on the recorded history data will be given and discussed in detail later in Section V. Let define the following vectors for this finite-state trajectory of length $L$ as
\begin{align}
\begin{array}{l}
R_L^\theta  = {[R({x_1}), \ldots ,R({x_L})]^T}\\
\,f_L^\theta  = {[{f_\theta }({x_1}), \ldots ,{f_\theta }({x_L})]^T}\\
\,\,\,{{\bar \delta }_L} = {[\delta (1), \ldots ,\delta (L)]^T}
\end{array} \label{eq:25}
\end{align}
and covariance vector and matrices are given by
%\begin{align}\nonumber
%K(x,X_L^{^\theta }) &= {[{k_0}({x_1},x), \ldots ,{k_0}({x_L},x)]^T},\\ %\nonumber
%K_L^\theta  &= {[K({x_1},X_L^{^\theta }), \ldots ,K({x_L},X_L^{^\theta %})]^T},\\
%{\sum _L} &= diag({w^2}, \ldots ,{w^2}), \label{eq:26}
%\end{align}
\begin{align}\nonumber
K(x,X_L^\theta ) &= [{k_0}\big((x_1,\theta),(x,\theta_1)\big), \dots ,{k_0}\big((x_L,\theta),(x,\theta_1)\big),
..., \nonumber \\
&\quad \quad {k_0}\big((x_1,\theta),(x,\theta_m)\big), \dots ,{k_0}\big((x_L,\theta),(x,\theta_m)\big)]_{,}^{T} \nonumber \\
K_L^\theta  &= {[K({x_1},X_L^{^\theta }), \ldots ,K({x_L},X_L^{^\theta })]_{,}^{T}} \nonumber  \\
{\sum}_{L} &= diag({w^2}, \ldots ,{w^2}), \label{eq:26}
\end{align}
where $\theta_1$ and $\theta_m$ are reward parameters for which their fitness is tested before. Incorporating tasks indexed or preview information also in the coaviance function allows transferring learning to similar contexts and finding a good prior for unseen contexts without relearning the GP.
Based on (25)-(26), one has
\begin{align}
\left( {\begin{array}{*{20}{c}}
{{{\bar \delta }_L}}\\
{f_L^\theta }
\end{array}} \right) \sim  {\cal N}\{ \left( {\begin{array}{*{20}{c}}
0\\
{m_0}(x,\theta)
\end{array}} \right),\left( {\begin{array}{*{20}{c}}
{{\sum _L}}&0\\
0&{K_L^\theta }
\end{array}} \right)\} \label{eq:27}
\end{align}

Using (24), for a finite state trajectory of length $L$, one has
\begin{align}
R_{L - 1}^\theta  = {H_L}f_L^\theta  + {\bar \delta _{L - 1}} \label{eq:28}
\end{align}
where
\begin{align}
{H_L} = {\left[ {\begin{array}{*{20}{c}}
1&{ - 1}&0& \ldots &0\\
0&1&{ - 1}& \ldots &0\\
 \vdots &{}&{}&{}& \vdots \\
0&0& \ldots &1&{ - 1}
\end{array}} \right]_{L - 1 \times L}} \label{eq:29}
\end{align}

Based on standard results on jointly Gaussian random variable, one has
\begin{align}
\Scale[0.8]{\left( {\begin{array}{*{20}{c}}
{R_{L - 1}^\theta }\\
{{f_\theta }(x)}
\end{array}} \right) \sim  {\cal N}\{ \left( {\begin{array}{*{20}{c}}
{H_L{m_0}(x,\theta)}\\
{{m_0}(x,\theta)}
\end{array}} \right),\left( {\begin{array}{*{20}{c}}
{{H_L}K_L^\theta H_L^T + {\sum _{L - 1}}}&{{H_L}K(x,X_L^\theta )}\\
{K{{(x,X_L^\theta )}^T}H_L^T}&{{k_0}\big((x,\theta),(x',\theta')\big)}
\end{array}} \right)\}} \label{eq:30}
\end{align}

Using (6), the posterior distribution of the fitness function ${f_\theta }(x(t))$ at state  $x(t)$, conditioned on observed integral meta-reward values $R_{t - 1}^\theta $  is given by
\begin{align}
({f_\theta }(x(t))|R_{t - 1}^\theta ) \sim  {\cal N}(\nu _t^\theta (x), p_t^\theta (x)) \label{eq:31}
\end{align}
where
\begin{align}
\begin{array}{l}
\,\,\,\,\,\,\,\nu _t^\theta (x) = {m_0}(x,\theta) + K{(x,X_t^{^\theta })^T}\alpha _t^\theta \\
p_t^\theta (x) = {k_0}\big((x,\theta),(x,\theta)\big) - K{(x,X_t^{^\theta })^T}C_t^\theta \,K(x,X_t^{^\theta })
\end{array} \label{eq:32}
\end{align}
with
\begin{align}
\begin{array}{l}
\alpha _t^\theta  = R{_{t - 1}^\theta}^T{({H_t}K_t^\theta H_t^T + {\sum _{t - 1}})^{ - 1}}{H_t}\\
\,C_t^\theta  = H_t^T{({H_t}K_t^\theta H_t^T + {\sum _{t - 1}})^{ - 1}}{H_t}
\end{array} \label{eq:33}
\end{align}

Based on (22), the following difference error is used as a surprise signal to detect changes in the fitness.
\begin{align}
{\cal S}{\cal P}(t) = {\nu _t}(x) - {\nu _{t + T}}(x) - R({x_t}) \label{eq:34}
\end{align}
where ${\nu _t}(x)$ is the mean of the GP and ${\cal S}{\cal P}(t)$ is the surprise signal at the time $t$. Note that after the GP learns the fitness, the surprise signal will be small, as the GP is learned to assure that (22) is satisfied. However, once the robustness degree changes due to a change in the situation, the surprise signal will increase, and if the average of the surprise signal over a horizon is bigger than a threshold, i.e.,
if
\begin{align}
\int_t^{t + \Delta } {{\cal S}{\cal P}(\tau )} \,d\tau  \ge \beta \label{eq:35}
\end{align}
for some threshold $\beta > 0$
then, the fitness function for the new situation will be inferred from the learned GP to realize if the change is due to factors that can be previewed or factors that cannot be observed directly and their effects can only be observed through unexpected system behavior. This can be checked by first inferring a fitness function from the GP and then monitoring \eqref{eq:35} based on the new inferred fitness function. If this value exceeds a threshold, it will indicate a change in the context that requires re-learning or fine-tuning the GP. A Bayesian optimization will be solved (using the current learned GP for the former case and the new learned GP for the latter case), as detailed later, to find a new set of hyperparameters that assure the expected value of the fitness function over a set of interest is below a threshold.  

The metacognitive layer does not adapt the hyperparameters all the time and only adapts them when two conditions are satisfied: 1) an event indicating a change is triggered, i.e., if (35) is satisfied for the inferred fitness function, 2) if inferred fitness is below a threshold, i.e., it does not indicate future safety and liveness. 

Note that while the second requirement can be ignored and the metacognitive layer can find a new set of hyperparameters as long as the first condition is satisfied, the second requirement can reduce the frequency of adaptation of metacongitive controller. To monitor the second requirement for detecting a threat that requires adapting the hyperparameters, inspired by \cite{Engel:36}, a KL divergence metric is defined to measure the similarity that GP learned for the current fitness and a base GP. The base GP can be obtained based on the knowledge of the constraints and STL specifications to assure the minimum safety of the system. Note that constructing the base GP only requires the knowledge of the STL specifications and is independent of the system dynamics, the situation, or the control objectives. A library of safe GPs can also be constructed as based GP and the previous learned GPs for other circumstances can be added. If the fitness remains close to any of the GPs in this library, this indicates that the system safety is still not in danger. If not, it is highly likely that the system's safety is in danger of being violated in the near future.

Since the covariance function $K$ corresponds to a (possibly infinite-dimensional) feature space, the Gaussian process can be viewed as a Gaussian distribution in the feature space. To show this, let $\phi (x)$ be the feature space representation of the covariance function so that $K_x(x,x) = \phi {(x)^T}\phi (x)$ (the same can be defined for $K_{\theta}(\theta,\theta)$). Then, we use the notation 
\begin{align}
{f_\theta }(x) = G{P_K}(\alpha _t^\theta ,C_t^\theta ) \label{eq:36}
\end{align}
to denote the GP with the corresponding covariance function $K$ and parameters  $\alpha _t^\theta$ and $C_t^\theta$.  We define the GP for the base fitness as
\begin{align}
{f_b}(x) = G{P_K}(\alpha _t^b,C_t^b) \label{eq:37}
\end{align}
  
\vspace{-0.3cm}  
  
\smallskip

\noindent
{\bf Lemma 5 \cite{McIntire:37}.}  Let the base GP and the GP for the hyperparameter vector  $\theta$ share the same inducing inputs, i.e., $X_L^b = X_L^{^\theta } = [{x_1},...,{x_L}]$, and the same covariance function $K$.  Let $K(X_L^{^\theta },X_L^b) = {Q^{ - 1}}$. Then, the KL divergence between two dynamic GPs ${f_\theta }(x)$  and ${f_b}(x)$ is given by 
\begin{align}
\begin{array}{l}
{D_{KL}}({f_\theta }(x)||{f_b}(x)) = {D_{KL}}(G{P_K}(\alpha _t^\theta ,C_t^\theta )||G{P_K}(\alpha _t^b,C_t^b))\\
\,\,\,\,\,\,\,\,\,\,\,\,\,\,\,\,\,\,\,\,\,\,\,\,\,\,\,\,\,\,\,\, = {(\alpha _t^\theta  - \alpha _t^b)^T}V(\alpha _t^\theta  - \alpha _t^b) + W
\end{array} \label{eq:38}
\end{align}
where 
\vspace{-0.2cm}
\begin{align}
V = {(Q + C_t^\theta )^{ - 1}} \nonumber
\end{align}
and
\begin{align}
W = {\rm{Tr[}}(Q + C_t^b)V - I] - \log {\rm{[}}(Q + C_t^b)V] \nonumber
\end{align}

\smallskip

\noindent
{\bf Remark 4.} Since the base fitness function is learned offline based on the minimum acceptable degree of robustness, one can select many inducing points for the base fitness function and retain only a subset of them in the expressions of the posterior mean and kernel functions that increase the similarity of the inducing inputs for both GPs. To use the KL divergence metric (38), one can use the fact that $K(x,x) = \phi {(x)^T}\phi (x)$,  and so $K(x,{x_1}) = K(x,{x_2})K({x_1},{x_2})$, to shift the inducing points of the base fitness function to those of the learned GP. 

Let ${f_b}(x) = [{f_{{b_1}}}(x),...,{f_{{b_M}}}(x)]$ be the stack of base GPs. After a change, if $\mathop {\min }\limits_i {D_{KL}}({f_\theta }(x)||{f_{{b_i}}}(x)) \le \varpi$, with  ${f_b}(x)$ as the fitness before the change, then this indicates that the system is still safe despite the change. Therefore, the monitor now triggers an event that requires adaptation of the hyperparameters if the following STL condition is violated
\begin{align}
\begin{array}{l}
\varphi  = \square {\rm{ (}}\underbrace {{\rm{(}}\int_t^{t + \Delta } {{\cal S}{\cal P}(\tau )} \,d\tau  > \beta {\rm{)}}}_{{\varphi _1}} \wedge \\
\,\,\,\,\,\,\,\,\,\,\,\,\,\,\,\,\,\underbrace {\mathop {\min }\limits_i ({D_{KL}}({f_\theta }(x)||{f_{{b_i}}}(x)) > \varpi )}_{{\varphi _2}})
\end{array} \label{eq:39}
\end{align}

\subsection{Metacognitive Control}

In this section, after the STL specification (39) is violated, a metacognitive controller is presented to find a new set of hyperparameters to guarantee that the fitness is improved, i.e., $\mathop {\max }\limits_i {D_{KL}}({f_\theta }(x)||{f_{{b_i}}}(x)) \le \varpi$  with the minimum sacrifice on the performance. That is, it is desired to assure safety while achieving as much performance as possible close to the optimal performance found prior to the threat. Note that if the base GPs are not available, one can find hyperparametrs that assure $E_x f_{\theta}(x) \le \varpi$ and use empirical risk minimization to approximate the expectation. 

Let ${\theta ^*}$  be the optimal hyperparameter found to optimize the performance prior to the change. The metacognitive layer then performs the following optimization to find the reward parameters (the preview information incorporated in $\theta$ are not under control). 

\begin{align}
\begin{array}{l}
\,\,\,\,\,\,\,\,\min \,\,\left\| {\theta  - {\theta ^*}} \right\|\\
s.t.\,\,\,\mathop {\min }\limits_i {D_{KL}}({f_\theta }(x)||{f_{{b_i}}}(x)) \le \varpi 
\end{array} \label{eq:40}
\end{align}

To solve this optimization, one can define the survival score function in the metacognitive layer as  
\begin{align}
{\cal H}(\theta ) = ||\theta  - {\theta ^*}|| + \log (1 + \frac{1}{{\varpi  - \mathop {\min }\limits_i {D_{KL}}({f_\theta }(x)||{f_{{b_i}}}(x))}}) \label{eq:41}
\end{align}

In this paper, safe Bayesian optimization (SBO) \cite{Berkenkamp:38} is used to find the optimal hyperparameters that optimize the survival score function in (41). SBO algorithm provided in Algorithm 1 guarantees safety using only evaluation of hyperparameters that lead to a safe score threshold with high confidence. This threshold is chosen as a value below which we do not want the system fitness to fall, which can be an indication of great risk of violation of specifications. SBO is a sample efficient optimization algorithm that requires only a few evaluations of the hyper performance or survival score function to find the optimal hyperparameters. While the safe set of hyperparameters is not known in the beginning, it is estimated after each function evaluation of SBO. In fact, at each iteration, SBO tries to not only look for the global maximum within the safe set known up to now (i.e., exploitation), but also increase the set of hyperparameters that safe to search within (i.e., exploration) as described in Algorithm 1. More specifically, the SBO builds a surrogate model ${\cal P}$  that maps the hyperparameters to the survival score ${\cal H}(\theta )$ in the metacognitive layer and expresses as 
\begin{align}
{\cal P}: {\cal D}^{'} \to \mathbb{R} \label{eq:42}
\end{align}
where ${{\cal D}^{'}}$ denotes the bounded domain of hyperparameters. 

Note that the latent function ${\cal P}$  in (42) is unknown, as the dynamics are not known. ${\cal P}$  can be sampled by running the system and evaluating the survival score function ${\cal H}(\theta )$ from the recorded data. These samples of ${\cal P}$ are generally uncertain because of noisy data and the score function is typically nonconvex, and no gradients are easily available. 

Now, we used GP to perform a non-parametric regression for the latent function ${\cal P}$ \cite{Rasmussen:32} and then SBO is used to optimize the survival score and determine the optimum values of hyperparameters. For the non-parametric regression, a prior mean function is defined as ${\mu _0}(\theta )$ to encode prior knowledge about the survival score function  ${\cal P}(\theta )$, and a covariance function $k(\theta ,{\theta ^{'}})$ is defined to encode the covariance of any two function values, and ${\cal P}(\theta )$ and ${\cal P}(\theta^{'})$ are leveraged to model the the mean estimates uncertainties. One can predict the survival score of the system corresponding to the hyperparameter vector $\theta $ by calculating its mean ${\mu _k}(\theta )$ and covariance ${\sigma _k}^2(\theta )$ over a set of $k$ observations $\{ {\theta _{1:k}},\,\,{{\cal P}_{1:k}}\} $ and given by   ${\cal P}(\theta ) \sim {\cal N}({\mu _k}(\theta ), {\sigma _k}^2(\theta ))$.    

Based on the predicted score function, i.e., ${\cal P}(\theta ) \sim {\cal N}({\mu _k}(\theta ), {\sigma _k}^2(\theta ))$, the lower and upper bound of the confidence interval at iteration $k$ is given as
\begin{align}
\left\{ \begin{array}{l}
{{\bar m}_k}(\theta ) = {\mu _{k - 1}}(\theta ) - {\beta _k}{\sigma _{k - 1}}(\theta )\\
{M_k}(\theta ) = {\mu _{k - 1}}(\theta ) + {\beta _k}{\sigma _{k - 1}}(\theta )
\end{array} \right. \label{eq:43}
\end{align}
where ${\beta _k} > 0$ denotes a scalar factor that defines the desired confidence interval. Based on (43), the safe set of all the hyperparameters  $\theta$ that lead to survival score values above the threshold ${{\cal P}_{\min }}$ is given by 
\begin{align}
{{\cal S}_k} \leftarrow \{ \theta  \in {D^{'}}|{\bar m_k} \ge {{\cal P}_{\min }}\} \label{eq:44}
\end{align}
Then, a set of potential maximizers is defined as \cite{Berkenkamp:38}
\begin{align}
{{\cal T}_k} \leftarrow \{ \theta  \in {{\cal S}_k}|{M_k}(\theta ) \ge \mathop {\max }\limits_{{\theta ^{'}}} {\bar m_k}(\theta )\} \label{eq:45}
\end{align}
which contains the set of all safe hyperparameters that yield an upper confidence interval ${M_k}(\theta )$ above the best safe lower bound. In order to define, a set of potential expanders which quantifies whether a new set of hyperparameters can be classified as safe for a new observation, an optimistic characteristic function for expanders is given as
\begin{align}
{g_k}(\theta ) = \{ {\theta ^{'}} \in {D^{'}}\backslash {{\cal S}_k}|{\bar m_{k,(\theta ,{M_k}(\theta ))}}({\theta ^{'}}) \ge {{\cal P}_{\min }}\} \label{eq:46}
\end{align}
where ${\bar m_{k,(\theta ,{M_k}(\theta ))}}$ is the CP's lower bound based on prior data and $(\theta ,{M_k}(\theta ))$ is a data point with a noiseless measurement of the upper confidence bound. Based on (46), it can be decided that how many of previously unsafe points can be now classified as safe according to (44), assuming that ${M_k}(\theta )$ is measured while evaluating ${\cal P}(\theta )$. The characteristic function is positive if there is a non-negligible chance that the new data point expands the safe set. Therefore, the set of possible expanders is expressed by
\begin{align}
{{\cal G}_k} \leftarrow \{ \theta  \in {{\cal S}_k}|{g_k}(\theta ) > 0\} \label{eq:47}
\end{align}
Then, a new set of hyperparameters is selected to evaluate the performance on the real system by selecting a set of hyperparameters for which we are the most uncertain from the union of the sets ${{\cal G}_k}$  and ${M_k}$, i.e., at iteration $k$ the score function is evaluated at ${\theta _k}$ 
\begin{align}
{\theta _k} \leftarrow \arg {\max _{\theta  \in \{ {{\cal G}_k} \cup {{\cal T}_k}\} }}({M_k}(\theta ) - {\bar m_k}(\theta )) \label{eq:48}
\end{align}

\begin{algorithm}[!ht]
\caption{SBO for Metacognitive Control}
%\begin{algorithmic}
\begin{enumerate}
\item [1:] \textbf{procedure} 
\item [2:] Initialize $GP$  with  $({\theta _0},{\cal P}({\theta _0}))$.
\item [3:] \textbf{for} $k{\rm{ }} = 1, \ldots $ \textbf{for}
\item [4:] ${{\cal S}_k} \leftarrow \{ \theta  \in {D^{'}}|{\bar m_k} \ge {{\cal P}_{\min }}\} $ where ${\bar m_k}(\theta ) = {\mu _{k - 1}}(\theta ) - {\beta _k}{\sigma _{k - 1}}(\theta )$ .
\item [5:] ${{\cal T}_k} \leftarrow \{ \theta  \in {{\cal S}_k}|{M_k}(\theta ) \ge \mathop {\max }\limits_{{\theta ^{'}}} {\bar m_k}(\theta )\}$ where ${M_k}(\theta ) = {\mu _{k - 1}}(\theta ) + {\beta _k}{\sigma _{k - 1}}(\theta )$.
\item [6:] ${{\cal G}_k} \leftarrow \{ \theta  \in {{\cal S}_k}|{g_k}(\theta ) > 0\}$ where ${g_k}(\theta ) = \{ {\theta ^{'}} \in {D^{'}}\backslash {{\cal S}_k}|{m_{k,(\theta ,{M_k}(\theta ))}}({\theta ^{'}}) \ge {{\cal P}_{\min }}\}$.
\item [7:] ${\theta _k} \leftarrow \arg {\max _{\theta  \in \{ {{\cal G}_k} \cup {{\cal T}_k}\} }}({M_k}(\theta ) - {\bar m_k}(\theta ))$.
\item [8:] Obtain measurement ${\cal P}({\theta _k})$.
\item [9:] Update GP with $({\theta _k},{\cal P}({\theta _k}))$.
\item [10:] \textbf{end for}
\item [11:] \textbf{end procedure}
\end{enumerate}
%\end{algorithmic}
\end{algorithm}

The evaluation approach in (48) works well for expanding the safe set \cite{Berkenkamp:38}, with a trade-off between exploration and exploitation. The most uncertain parameters are usually located on the boundary of the safe set, which can be used for efficient exploration. An estimate of the best currently known set of hyperparameters is obtained from  $\mathop {\arg \max }\limits_{} \,\,{\bar m_k}(\theta ),\mathop {}\limits_{} \forall \theta  \in {{\cal S}_k}$ which corresponds to the point that achieves the best lower bound on the survival score.

\smallskip

\noindent
{\bf Remark 5.} It is shown in \cite{Berberich:39} and \cite{Willems:40} that, given a persistently exciting input, a single rich measured trajectory can be used to characterize the entire system trajectories. That is, having a single rich trajectory of an unknown system, the trajectory for a given sequence of inputs and an initial condition can be constructed without even applying it to the system. This can be leveraged to learn the fitness function for a given control policy selected by the Bayesian optimization algorithm without actually applying it to the system. More specifically, after a change, to evaluate the fitness of hyperparameters, a rich trajectory of the system is first collected and then used to reconstruct the trajectory of the system with enough length for any hyperparameter under evaluation. This trajectory data can then be used to learn the GP for the hyperparameters without even applying its corresponding controller. This will allow us to even evaluate unsafe policies without even applying them to the system. Hence, for each set of hyperparameters, one can compute the fitness function from measured data without knowledge of the closed-loop system’s dynamics, and consequently, find the optimal hyperparameters that optimize the survival score function in (40). This resembles off-policy learning in RL.

\smallskip

\noindent
{\bf Remark 6.}  Note that SBO is derivative-free nd thus can optimize the hyperparameter vector $\theta $  even though a closed-form expression of ${\cal P}$ as a function of $\theta $ is not available. Moreover, SBO can be leveraged to find the optimal set of hyperparameters with as few evaluations of ${\cal P}$ as possible. This is crucial because evaluation of ${\cal P}$ for a set of hyperparameters can be costly and time-consuming, as it requires a closed-loop experiment.

\section{LOW-LEVEL RL-BASED CONTROL ARCHITECTURE}

In this section, a computational data-driven algorithm is first developed to find the new optimal control policy $u_\theta ^*(t)$, for the new hyperparameter vector $\theta$ based on recorded history. It is then shown that the proposed algorithm converges to the optimal control policy $u_\theta ^*(t)$ for all admissible hyperparameters $\theta$. 

Following the same reasoning from \cite{Jiang:34}, we are ready to give the following computational data-driven algorithm for finding the new optimal control policy, i.e., $u_\theta ^*(t)$, for a new hyperparameter vector $\theta$ based on the recorded history data.

%\small
\begin{algorithm}[!htp]
\caption{Low-level Off-policy RL-based Control}
%\begin{algorithmic}
\begin{enumerate}
%\item [1:] \textbf{procedure} 
\item [1:] Perform an experiment on the time interval $[{t_0},{t_l}]$ by applying fixed stabilizing control policy $u(t) + e$  to the system, where $e$  is the exploration noise, and then record $\{ {X_\theta }(t)\} $ and corresponding $\{ u(t)\} $  at $N \ge {l_1} + m \times {l_2}$  different sampling in the time interval $[{t_0},{t_l}]$, where ${X_\theta }(t) = {[{e_d}{(t)^T},{r_\theta }^T]^T}$, ${e_d}(t): = x(t) - {r_\theta }$.
\item [2:] For a new parameter of $\theta $ where $\theta  \in \mathchar'26\mkern-10mu\lambda $, construct  ${\phi _1}({X_\theta }) \in {\mathbb{R}^{{l_1}}}$ and $\Phi ({X_\theta }) \in {\mathbb{R}^{{l_2}}}$ as suitable basis function vectors. Compute  ${\Xi _k}(\theta )$ and ${\Theta _k}(\theta )$ for a new set of parameters $\theta $ based on the recorded history data
 \begin{align}
 &\,\,\,{\Xi _k}(\theta ) =  - {I_{xx}}(\theta )\left[ {\begin{array}{*{20}{c}}
 {vec({{\bar Q}_\theta })}\\
 {vec({R_\theta })}
 \end{array}} \right] \label{eq:49} \\
 & \begin{array}{l}
  {\Theta _k}(\theta ) = [{\delta _{xx}}, - 2{{\bar I}_{xx}}(\theta )({I_n} \otimes \hat {\bar W}_k^T{R_\theta })\\
  \,\,\,\,\,\,\,\,\,\,\,\,\,\,\,\,\,\,\,\,\,\,\,\,\,\,\,\,\,\,\,\,\,\,\,\,\,\,\,\,\,\,\,\,\,\,\,\,\, - 2{I_{xu}}(\theta )({I_n} \otimes {R_\theta })]
  \end{array} \label{eq:50}
  \end{align} 
  where ${\bar Q_\theta }: = diag({Q_\theta },0)$ and
    \begin{align}
    &  \quad\,\, {\delta _{xx}} = [{e^{ - \gamma T}}({\phi _1}({X_\theta }({t_1})) - {\phi _1}({X_\theta }({t_0}))), \nonumber \\
    & \quad\,\, \quad\,\, \quad\,\, {e^{ - \gamma T}}({\phi _1}({X_\theta }({t_2})) - {\phi _1}({X_\theta }({t_1}))), \nonumber \\
    & \quad\,\, \quad\,\, \quad\,\, {e^{ - \gamma T}}({\phi _1}({X_\theta }({t_l})) - {\phi _1}({X_\theta }({t_{l - 1}}))){]^T} \label{eq:51} \\
    &  \begin{array}{l}
       {{\bar I}_{xx}}(\theta ) = [\int_{{t_0}}^{{t_1}} {{e^{ - \gamma (\tau  - t)}}(\Phi ({X_\theta }) \otimes \Phi ({X_\theta }))d\tau } ,\\
       \,\,\,\,\,\,\,\,\,\,\,\,\,\,\,\,\,\,\,\,\,\,\,\,\,\,\int_{{t_1}}^{{t_2}} {{e^{ - \gamma (\tau  - t)}}(\Phi ({X_\theta }) \otimes \Phi ({X_\theta }))d\tau } ,...,\\
       \,\,\,\,\,\,\,\,\,\,\,\,\,\,\,\,\,\,\,\,\,\,\,\,\,\,\int_{{t_{l - 1}}}^{{t_l}} {{e^{ - \gamma (\tau  - t)}}(\Phi ({X_\theta }) \otimes \Phi ({X_\theta }))d\tau } {]^T}
       \end{array} \label{eq:52}
       \end{align}
 \begin{align}
  &      \Scale[0.88]{ {I_{xx}}(\theta ) = 
{\left[ \begin{array}{l}
\,\,\,\,\,\,\,\,\,\,\,\left[ {\int_{{t_0}}^{{t_1}} {{e^{ - \gamma (\tau  - t)}}{X_\theta } \otimes {X_\theta }d\tau } ,} \right.\\
\,\,\,\,\,\,\,\,\,\,\,\,\int_{{t_1}}^{{t_2}} {{e^{ - \gamma (\tau  - t)}}{X_\theta } \otimes {X_\theta }d\tau ,} \\
\left. {\,\,\,\,\,\,\,\,\,\,\,\,\,\,\,...,\int_{{t_{l - 1}}}^{{t_l}} {{e^{ - \gamma (\tau  - t)}}{X_\theta } \otimes {X_\theta }d\tau } } \right]\\
\left[ {\int_{{t_0}}^{{t_1}} {{e^{ - \gamma (\tau  - t)}}(\hat {\bar W}_k^T\Phi ({X_\theta }) \otimes \hat {\bar W}_k^T\Phi ({X_\theta }))d\tau } ,} \right.\\
\int_{{t_1}}^{{t_2}} {{e^{ - \gamma (\tau  - t)}}(\hat {\bar W}_k^T\Phi ({X_\theta }) \otimes \hat {\bar W}_k^T\Phi ({X_\theta }))d\tau } ,\\
{\mkern 1mu} \left. {...,\int_{{t_{l - 1}}}^{{t_l}} {{e^{ - \gamma (\tau  - t)}}(\hat {\bar W}_k^T\Phi ({X_\theta }) \otimes \hat {\bar W}_k^T\Phi ({X_\theta }))d\tau } } \right]
\end{array} \right]^T}   }   
 \label{eq:53} \\
           &\begin{array}{l}
           {I_{xu}}(\theta ) = [\int_{{t_0}}^{{t_1}} {{e^{ - \gamma (\tau  - t)}}(\Phi ({X_\theta }) \otimes u)d\tau },\\
           \quad \quad\quad\quad\,\,\,\int_{{t_1}}^{{t_2}} {{e^{ - \gamma (\tau  - t)}}(\Phi ({X_\theta }) \otimes u)d\tau } ,\\
           \quad \quad\quad\quad\,\,\,\int_{{t_{l - 1}}}^{{t_l}} {{e^{ - \gamma (\tau  - t)}}(\Phi ({X_\theta }) \otimes u)d\tau } {]^T}
           \end{array} \label{eq:54}
           \end{align} 
\item [3:] Solve $\hat W_k^V \in {\mathbb{R}^{{l_1}}}$ and  $\hat {\bar W}_{k + 1}^{} \in {\mathbb{R}^{{l_2} \times m}}$ from (55)
 \begin{align}
  \left[ {\begin{array}{*{20}{c}}
  {\hat W_k^V}\\
  {vec(\hat {\bar W}_{k + 1}^T)}
  \end{array}} \right] = {(\Theta _k^T(\theta ){\Theta _k}(\theta ))^{ - 1}}\Theta _k^T(\theta ){\Xi _k}(\theta ) \label{eq:55}
  \end{align} 
and update the value function and control policy
 \begin{align}
  \hat V_k^\theta ({X_\theta }): &= {(\hat W_k^V)^T}{\phi _1}({X_\theta }) \label{eq:56} \\
  u_\theta ^{k + 1}(t): &= \hat {\bar W}_{k + 1}^T\Phi ({X_\theta }) \label{eq:57}
    \end{align} 
\item [4:] Let $k \leftarrow k + 1$, and go to Step 3 until  $\left\| {\hat W_k^V - \hat W_{k - 1}^V} \right\| \le \varepsilon$ for $k{\rm{ }} \ge {\rm{ }}1$, where the constant $\varepsilon {\rm{ }} > {\rm{ }}0$ is a predefined threshold.
\item [5:] Use $u_\theta ^*(t): = u_\theta ^{k + 1}(t)$ and ${V^{\theta}} ^*({X_{\theta} }): = \hat V_k^{\theta} ({X_{\theta} })$ as the approximated optimal control policy and its approximated optimal value function corresponding to a new set of $\theta$, respectively.
%\item [7:] \textbf{end procedure}
\end{enumerate}
%\end{algorithmic}
\end{algorithm}

\smallskip

\noindent
{\bf Remark 7.} Note that Algorithm 2 does not rely on the dynamics of the system. Note also that, inspired by the off-policy algorithm in \cite{Jiang:34}, Algorithm 2 has two separate phases. In the first phase, i.e., Step 2, a fixed  exploratory control policy $u$ is deployed, and the system's states are recorded over the interval $[{t_0},{t_l}]$. In the second phase, i.e., Steps 3-7, then, without the requirement of knowing any knowledge about the system's dynamics, the information collected in the first phase is used as many times as required to learn a sequence of updated control policies converging to ${u^ * }$. 

The following theorem shows that Algorithm 2 converges to the optimal control policy $u_\theta ^*(t)$ for all admissible hyperparameter vectors $\theta$.

\smallskip

\noindent
{\bf Theorem 2 (Convergence of Algorithm 2).} Let the new hyperparameter vector $\theta $ be admissible. Using the fixed stabilizing control policy $u(t)$, when $N \ge {l_1} + m \times {l_2}$, $u_\theta ^{k + 1}(t)$ obtained from solving (55) in the off-policy Algorithm 2, converges to the optimal control policy $u_\theta ^*(t)$, $\forall \theta  \in \mathchar'26\mkern-10mu\lambda$.\\

\smallskip

\noindent
 {\bf Proof.} $\forall \theta  \in \mathchar'26\mkern-10mu\lambda $, set  ${Q_\theta }$, ${R_\theta }$, and ${r_\theta }$. Using (49)-(54), it follows from (55) that $\hat W_k^V \in {\mathbb{R}^{{l_1}}}$ and  $\hat {\bar W}_{k + 1}^{} \in {\mathbb{R}^{{l_2} \times m}}$ satisfy the following Bellman equation. 
 \begin{align}
 \begin{array}{l}
 {e^{ - \gamma t}}\hat W_{k}^T({\phi _1}({X_\theta }(t + \delta t)) - {\phi _1}({X_\theta }(t)))\\
  =  - \int_t^{t + \delta t} {{e^{ - \gamma (\tau  - t)}}({{({X_\theta })}^T}{{\bar Q}_\theta }({X_\theta })) + {u_k}^T{R_\theta }{u_k})d\tau } \\
 \,\,\,\, + \int_t^{t + \delta t} {{e^{ - \gamma (\tau  - t)}}( - 2(\hat {\bar W}_{k + 1}^T\Phi ({X_\theta })){R_\theta }(u - {u_k})) d\tau } 
 \end{array} \label{eq:58}
 \end{align}  
 Now, let $W_{}^V \in {\mathbb{R}^{{l_1}}}$  and $\bar W_{}^{} \in {\mathbb{R}^{{l_2} \times m}}$ such that  ${u_k} = \bar W_{}^T\Phi ({X_\theta })$ and   
    \begin{align}
    \left[ {\begin{array}{*{20}{c}}
    {W_{}^V}\\
    {vec(\bar W_{}^T)}
    \end{array}} \right] = {(\Theta _k^T(\theta ){\Theta _k}(\theta ))^{ - 1}}\Theta _k^T(\theta ){\Xi _k}(\theta ) \label{eq:59}
    \end{align}   
Then, one immediately has $W_{}^V = \hat W_k^V$  and $vec(\bar W_{}^{}) = vec(\hat {\bar W}_{k + 1}^T)$. If the condition given in Step 2 is satisfied, i.e., the information is collected at $N \ge {l_1} + m \times {l_2}$ points, then ${[\hat W_k^V,\,\,vec(\hat {\bar W}_{k + 1}^T)]^T}$  has $N$  independent elements, and therefore the solution of least squares (LS) in (55) and (59) are equal and unique. That is, $\hat W_k^V = W_{}^V$  and $\hat {\bar W}_{k + 1}^{} = \bar W$. This completes the proof.  \hfill $\square$
     
 \section{SIMULATION RESULTS}
 The presented algorithm is validated for the steering control of an autonomous vehicle in the lane-changing scenario. Consider the following vehicle dynamics for steering control as \cite{Souza:41}
\begin{align}
\dot x(t) = Ax(t) + Bu(t) \label{eq:60}
 \end{align}
\begin{align}
\begin{array}{l}
A = \left[ {\begin{array}{*{20}{c}}
0&{{v_T}}&{{v_T}}&0\\
0&0&0&1\\
0&0&{ - \frac{{{k_f} + {k_r}}}{{{m_T}{v_T}}}}&{ - \frac{{{m_T}{v_T} + \frac{{a{k_f} - b{k_r}}}{{{v_T}}}}}{{{m_T}{v_T}}}}\\
0&0&{ - \frac{{a{k_f} - b{k_r}}}{{{I_T}}}}&{ - \frac{{{a^2}{k_f} + {b^2}{k_r}}}{{{I_T}{v_T}}}}
\end{array}} \right],{\mkern 1mu} {\mkern 1mu} {\mkern 1mu} {\mkern 1mu} \\
{B^T} = \left[ {\begin{array}{*{20}{c}}
0&0&{\frac{{{k_f}}}{{{m_T}{v_T}}}}&{\frac{{a{k_f}}}{{{I_T}}}}
\end{array}} \right]
\end{array} \label{eq:61}
  \end{align}
 where $x(t) = {[\begin{array}{*{20}{c}}
 {{x_1}}&{{x_2}}&{{x_3}}
 \end{array}\,\,\,\,{x_4}]^T} = {[\begin{array}{*{20}{c}}
 y&\psi &\alpha 
 \end{array}\,\,\,\,\psi ]^T}$. The state variables are defined as the lateral position of vehicle $y$, yaw angle  $\psi$, slip angle $\alpha$ and rate of change of yaw angle $\dot \psi$.  Moreover, $\delta$ represents the steering angle and acts as the control input, ${v_T}$ denotes the longitudinal speed of the vehicle, ${m_T}$ is the vehicle's total mass, ${I_T}$ is the inertia moment with respect to the mass center, ${k_f}$ and ${k_r}$ are the tire's stiffness parameters, and $a$ and $b$ denote the distance of the front and rear tires to the mass center. 
 
  The values of vehicle parameters used in the simulation are provided in Table I. For the validation of the presented algorithm, lane-changing scenario for a two-lane highway is considered as shown in Fig.2. In this simulation following STL constraint is considered, the vehicle state is subject to the desired specification on the offset from the centerline, i.e.,  $\varphi  = \square {\rm{ (}}\left| {{x_1} - r} \right| < 1)$  with $r$ as the center lane trajectory (act as setpoint value). In the simulation, setpoint value is selected as $r = 1$ for $t < 4s$, otherwise $r = 3$. Fig.3 shows the result for lane-changing scenario with a fixed values of hyperparameters and without any change in system dynamics. The control policy or steering angle in (60) is evaluated based on off-policy RL algorithm in \cite{Jiang:34} with hyperparameter values as $Q = diag(10,10,10,10)$  and $R = 2$. Then, in Fig. 4, we consider the change in the system dynamics after t=4s and $t=4s$ it becomes   
   \begin{align}
  \dot x(t) = (A + \Delta A)x(t) + (B + \Delta B)u(t) \label{eq:62}
   \end{align}
with   
    \begin{align}
   \begin{array}{l}
   \Delta A = \left[ {\begin{array}{*{20}{c}}
   0&{\Delta {v_T}}&{\Delta {v_T}}&0\\
   0&0&0&0\\
   0&0&{ - \frac{{{k_f} + {k_r}}}{{{m_T}\Delta {v_T}}}}&{ - \frac{{{m_T}\Delta {v_T} + \frac{{a{k_f} - b{k_r}}}{{\Delta {v_T}}}}}{{{m_T}\Delta {v_T}}}}\\
   0&0&0&{ - \frac{{{a^2}{k_f} + {b^2}{k_r}}}{{{I_T}\Delta {v_T}}}}
   \end{array}} \right],{\mkern 1mu} {\mkern 1mu} {\mkern 1mu} {\mkern 1mu} \\
   \Delta {B^T} = {\left[ {\begin{array}{*{20}{c}}
   0&0&{\frac{{{k_f}}}{{{m_T}\Delta {v_T}}}}&0
   \end{array}} \right]^T}
   \end{array}\label{eq:63}
    \end{align}
    
  The vehicle parameter values for (63) are provided in Table I. Control input is evaluated based on off-policy RL algorithm in \cite{Jiang:34} with fixed hyperparameter values as $Q = diag(10,10,10,10)$ and $R = 2$. The result in Fig. 4 shows that that the vehicle starts wavering and goes out of the lane. That is, the vehicle violates the desired specification, i.e., $\varphi  = \square {\rm{ (}}\left| {{x_1} - r} \right| < 1)$ after the change in the dynamics. 
  
   \begin{table}[!t]
   \renewcommand{\arraystretch}{1.3}
   % if using array.sty, it might be a good idea to tweak the value of
  %  \extrarowheight as needed to properly center the text within the cells
   \caption{Vehicle Parameters}
   \label{Table1}
   \centering
   \begin{tabular}{|c|c|c|c|}
   \hline
   Parameter & Value & Parameter & Value \\
    \hline
   ${m_T}$ & $1300\,\,kg$ & ${k_f}$ & $91000\,\,N/rad$ \\
   \hline
  ${I_T}$ & $10000\,\,{m^2}.kg$ & ${k_r}$ & $91000\,\,N/rad$ \\
    \hline
   ${v_T}$ & $16\,\,\,m/s$ & $\beta $ & $2$ \\
    \hline
   $a$ & $1.6154\,\,m$ & $b$ & $1.8846\,\,m$ \\
   \hline
   \end{tabular}
   \end{table}  
    
        \begin{figure}[!t]
        \centering{\includegraphics[width=3.5in] {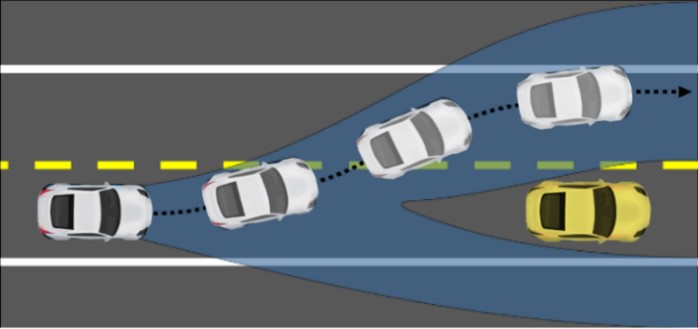}}
        \caption{Lane-changing scenario for the steering control of autonomous vehicle.}
        \label{fig2}
        \end{figure}
    
   \begin{figure}[!t]
   \centering{\includegraphics[width=3.5in] {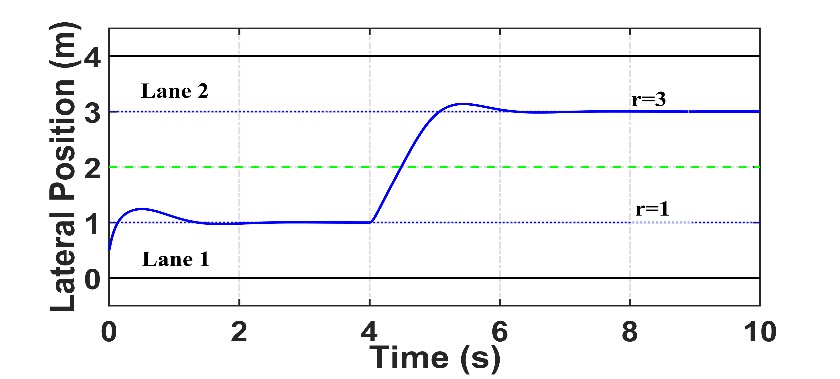}}
   \caption{Lane-changing with fixed value of the hyperparameter without any change in dynamics.}
   \label{fig3}
   \end{figure}
   
     \begin{figure}[!t]
     \centering{\includegraphics[width=3.5in] {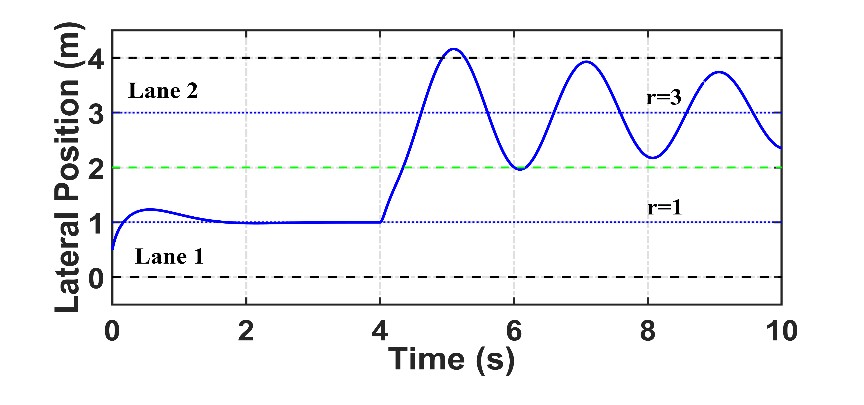}}
     \caption{Constraint violation during the lane-changing with the fixed value of hyperparameter and the change in dynamics.}
     \label{fig4}
     \end{figure}

       \begin{figure}[!t]
       \centering{\includegraphics[width=3.5in] {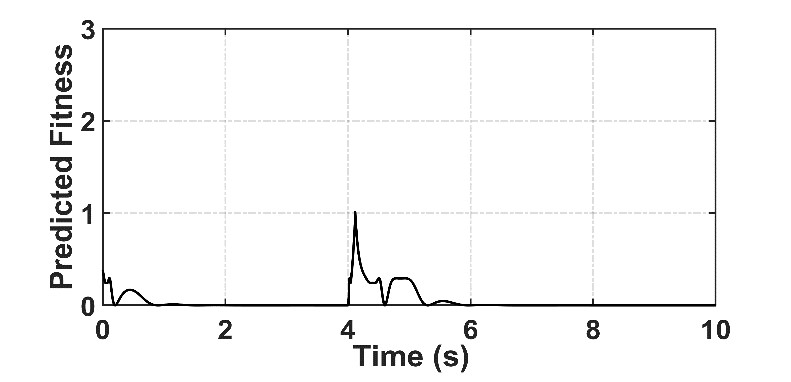}}
       \caption{Predicted fitness corresponding to vehicle trajectory in Fig. 3.}
       \label{fig5}
       \end{figure}

 Now, in order to implement the presented algorithm, first, the fitness function in (22) is learned as GP based on the temporal difference in (23). Note the fitness function ${f_\theta }(x)$ is learned offline and it is implemented for online monitoring and control in the meta-cognitive layer. Figs. 5 and 6 show the predicted fitness function based on the learned GP for the vehicle trajectories in Figs. 3 and 4. Based on the result in Figs. 6 and 7, one can see how the fitness value grows due to the operation of the system close or beyond the desired STL specification. The fitness value is used for meta-cognitive monitoring and intermittent evaluation of the meta-cognitive control layer. 
 
 Based on the meta-cognitive monitor in (39), Algorithm 1 is evaluated using the survival score function in (41) and determines the optimum hyperparameter to ensure the desired STL specification, i.e., $\varphi  = \square {\rm{ (}}\left| {{x_1} - r} \right| < 1)$.  Fig. 8 shows the vehicle trajectory with hyperparameter adaptation based on Algorithm 1 with the change in dynamics. The new optimum hyperparameter values are found to be  $Q = diag(96.11,1.2,1,1.5)$  and $R = 1$. Also, Fig. 9 shows how the predicted fitness value converges close to zero after the hyperparameter adaptation and the overall fitness value becomes constant as shown in Fig. 10. The presented algorithm is employed to learn control solutions with good enough performances while satisfying desired specifications and properties expressed in terms of STL. As shown in Figs. 8 and 9, based on meta-cognitive layer hyperparameters are adapted and lane-changing problem for autonomous vehicle is solved without violating any constraint.    
  
         \begin{figure}[!t]
         \centering{\includegraphics[width=3.5in] {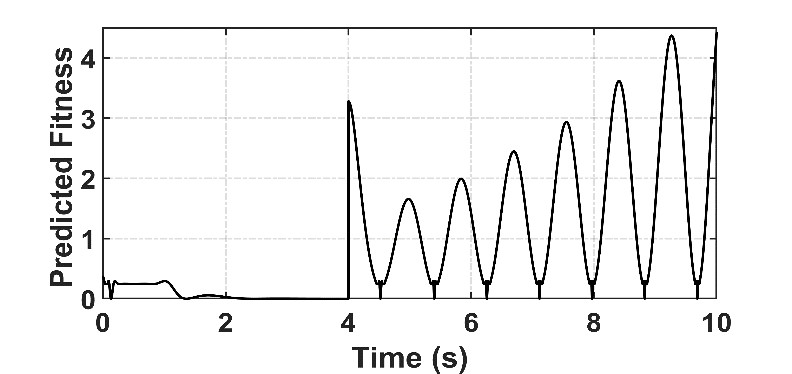}}
         \caption{Predicted fitness corresponding to vehicle trajectory in Fig. 4.}
         \label{fig6}
         \end{figure}

           \begin{figure}[!t]
           \centering{\includegraphics[width=3.5in] {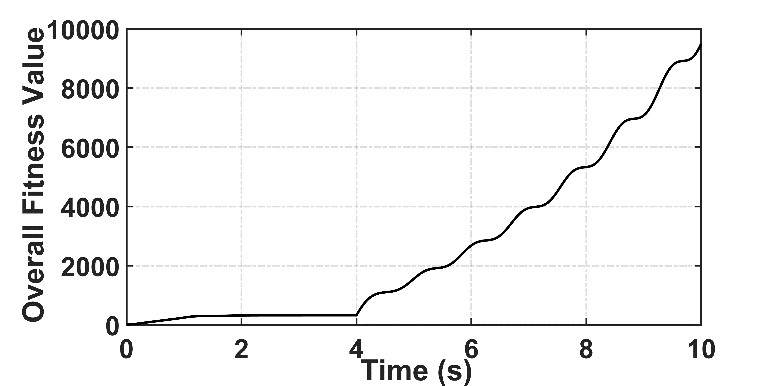}}
           \caption{Overall fitness value under desired STL constraint violation for the vehicle trajectory in Fig.4.}
           \label{fig7}
           \end{figure}
           
                      \begin{figure}[!t]
                      \centering{\includegraphics[width=3.5in] {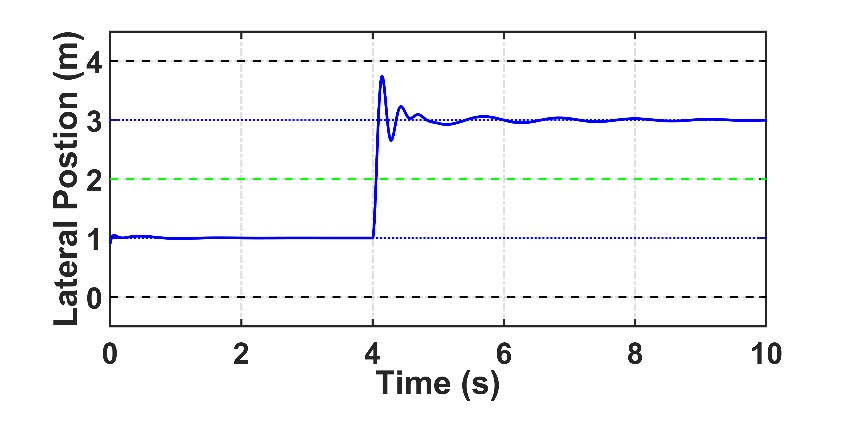}}
                      \caption{Vehicle trajectory with hyperparameter adaptation based on Algorithm 1 for the lane-changing scenario.}
                      \label{fig8}
                      \end{figure}
                      
                                 \begin{figure}[!t]
                                 \centering{\includegraphics[width=3.5in] {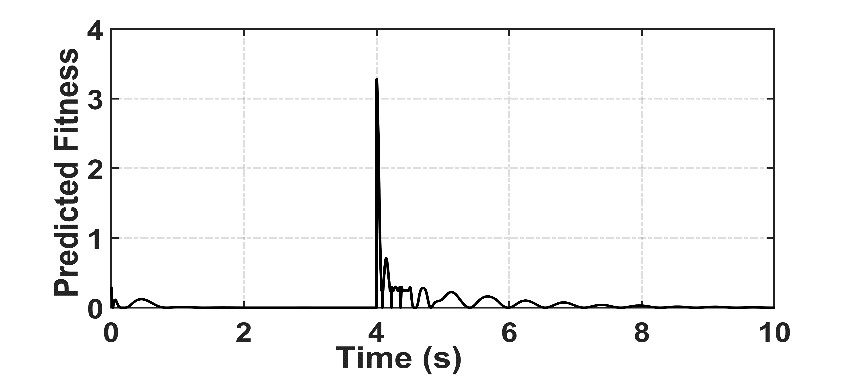}}
                                 \caption{Predicted fitness corresponding to vehicle trajectory in Fig. 8}
                                 \label{fig9}
                                 \end{figure}
                                 
                                        \begin{figure}[!t]
                                         \centering{\includegraphics[width=3.5in] {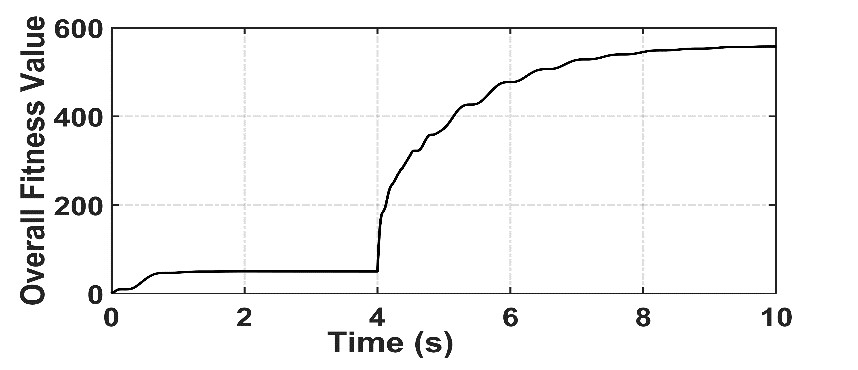}}
                                         \caption{Overall fitness value under desired STL constraint for adapted vehicle trajectory in Fig. 8.}
                                         \label{fig10}
                                         \end{figure}
 
\section{Conclusion}

An assured metacognitive RL-based autonomous control framework is presented to learn control solutions with good enough performances while satisfying desired specifications and properties expressed in terms of Signal temporal logic (STL). We discussed that the pre-specified reward functions cannot guarantee the satisfaction of the desired specified constraints and properties across all circumstances that an uncertain system might encounter. That is, the system either violates safety specifications or achieves no optimality and liveness specifications. In order to overcome this issue, learning what reward functions to choose to satisfy desired specifications and to achieve a good enough performance across a variety of circumstances, a metacognitive decision-making layer is presented in augmentation with the performance-driven layer.  More specifically, an adaptive reward function is presented in terms of its gains and adaptive reference trajectory (hyperparameters), and these hyperparameters are determined based on metacognitive monitor and control to assure the satisfaction of the desired STL safety and liveness specifications. The proposed approach separates learning the reward function that satisfies specifications from learning the control policy that maximizes the reward and thus allows us to evaluate as many hyperparameters as required using reused data collected from the system dynamics.

\end{document}